\documentclass[10pt,twocolumn,letterpaper]{article}

\usepackage[]{wacv}
\usepackage[table]{xcolor}
\usepackage[accsupp]{axessibility}

\usepackage{graphicx}
\usepackage{epsfig}
\usepackage{amsmath}
\usepackage{amssymb}
\usepackage{booktabs}
\usepackage{colortbl}
\usepackage{textcomp}
\usepackage{subcaption}
\usepackage[mode=buildnew]{standalone}
\usepackage{pgfplots} 
\usepackage{multirow}
\usepackage{svg}
\usepackage{lipsum}
\usepackage{pifont}
\usepackage{yfonts}
\usepackage{listings}
\usepackage{algorithm}
\usepackage{algpseudocode}
\usepackage{makecell}
\usepackage[T1]{fontenc}
\usepackage[normalem]{ulem}
\usepackage{nicematrix}
\usepackage{placeins}
\usepackage{circledsteps}
\usepackage{scalerel}

\newcommand{\VEC}[1]{\mathbf{#1}}          %
\newcommand{\VECG}[1]{\boldsymbol{#1}}     %

\newcommand{\putindex}[3]{\vtop{\hbox{\hspace{#3} $#1$}
            \hbox{\raise 6mm \hbox{$\scriptscriptstyle #2$}}}}

\newcommand{\gradx}[0]{\vtop{\hbox{\rm grad}
            \hbox{\raise 2.5mm \hbox{\rm \hspace{2mm} \footnotesize x}}}}

\newcommand{\grady}[0]{\vtop{\hbox{\rm grad}
            \hbox{\raise 2.5mm \hbox{\rm \hspace{2mm} \footnotesize y}}}}

\newcommand{\grad}[1]{\vtop{\hbox{\rm grad}
            \hbox{\raise 2.5mm \hbox{#1}}}}

\newcommand{\btb}{     \begin{tabbing}             }
\newcommand{\bte}{     \end{tabbing}               }

\definecolor{tu0}{rgb}{0.7451, 0.1176, 0.2353}

\definecolor{tu1}{rgb}{1.0000, 0.8039, 0.0000}
\definecolor{tu11}{rgb}{1.0000, 0.8627, 0.3020}
\definecolor{tu12}{rgb}{1.0000, 0.9020, 0.4980}
\definecolor{tu13}{rgb}{1.0000, 0.9412, 0.6980}
\definecolor{tu14}{rgb}{1.0000, 0.9608, 0.8000}

\definecolor{tu2}{rgb}{0.9804, 0.4314, 0.0000}
\definecolor{tu21}{rgb}{0.9882, 0.6039, 0.3020}
\definecolor{tu22}{rgb}{0.9882, 0.7137, 0.4980}
\definecolor{tu23}{rgb}{0.9922, 0.8275, 0.6980}
\definecolor{tu24}{rgb}{0.9961, 0.8863, 0.8000}

\definecolor{tu3}{rgb}{0.6902, 0.0000, 0.2745}
\definecolor{tu31}{rgb}{0.7529, 0.2000, 0.4196}
\definecolor{tu32}{rgb}{0.8431, 0.4980, 0.6353}
\definecolor{tu33}{rgb}{0.9216, 0.7490, 0.8196}
\definecolor{tu34}{rgb}{0.9529, 0.8510, 0.8902}

\definecolor{tu4}{rgb}{0.4863, 0.8039, 0.9020}
\definecolor{tu41}{rgb}{0.6431, 0.8627, 0.9333}
\definecolor{tu42}{rgb}{0.7412, 0.9020, 0.9490}
\definecolor{tu43}{rgb}{0.8431, 0.9412, 0.9686}
\definecolor{tu44}{rgb}{0.8980, 0.9608, 0.9804}

\definecolor{tu5}{rgb}{0.0000, 0.5020, 0.7059}
\definecolor{tu51}{rgb}{0.3020, 0.6510, 0.7961}
\definecolor{tu52}{rgb}{0.5490, 0.7765, 0.8667}
\definecolor{tu53}{rgb}{0.7490, 0.8745, 0.9255}
\definecolor{tu54}{rgb}{0.8510, 0.9255, 0.9569}

\definecolor{tu6}{rgb}{0.0000, 0.3255, 0.4549}
\definecolor{tu61}{rgb}{0.2510, 0.4941, 0.5922}
\definecolor{tu62}{rgb}{0.5490, 0.6941, 0.7529}
\definecolor{tu63}{rgb}{0.7490, 0.8314, 0.8627}
\definecolor{tu64}{rgb}{0.8510, 0.8980, 0.9176}

\definecolor{tu7}{rgb}{0.0314, 0.0314, 0.0314}
\definecolor{tu71}{rgb}{0.3725, 0.3725, 0.3725}
\definecolor{tu72}{rgb}{0.5882, 0.5882, 0.5882}
\definecolor{tu73}{rgb}{0.7529, 0.7529, 0.7529}
\definecolor{tu74}{rgb}{0.8667, 0.8667, 0.8667}

\definecolor{tu8}{rgb}{0.7765, 0.9333, 0.0000}
\definecolor{tu81}{rgb}{0.8431, 0.9529, 0.3020}
\definecolor{tu82}{rgb}{0.8863, 0.9647, 0.4980}
\definecolor{tu83}{rgb}{0.9333, 0.9804, 0.6980}
\definecolor{tu84}{rgb}{0.9569, 0.9882, 0.8000}

\definecolor{tu9}{rgb}{0.5373, 0.6431, 0.0000}
\definecolor{tu91}{rgb}{0.6784, 0.7490, 0.3020}
\definecolor{tu92}{rgb}{0.7686, 0.8196, 0.4980}
\definecolor{tu93}{rgb}{0.8588, 0.8941, 0.6980}
\definecolor{tu94}{rgb}{0.9059, 0.9294, 0.8000}

\definecolor{tu10}{rgb}{0.0000, 0.4431, 0.3373}
\definecolor{tu101}{rgb}{0.3020, 0.6118, 0.5373}
\definecolor{tu102}{rgb}{0.5490, 0.7490, 0.7020}
\definecolor{tu103}{rgb}{0.7490, 0.8588, 0.8353}
\definecolor{tu104}{rgb}{0.8549, 0.9176, 0.9059}

\definecolor{tu110}{rgb}{0.8000, 0.0000, 0.6000}
\definecolor{tu111}{rgb}{0.8706, 0.3490, 0.7412}
\definecolor{tu112}{rgb}{0.9216, 0.6000, 0.8392}
\definecolor{tu113}{rgb}{0.9608, 0.8000, 0.9216}
\definecolor{tu114}{rgb}{0.9804, 0.8980, 0.9608}

\definecolor{tu120}{rgb}{0.4627, 0.0000, 0.4627}
\definecolor{tu121}{rgb}{0.5961, 0.2510, 0.5961}
\definecolor{tu122}{rgb}{0.7294, 0.4980, 0.7294}
\definecolor{tu123}{rgb}{0.8392, 0.6980, 0.8392}
\definecolor{tu124}{rgb}{0.9216, 0.8510, 0.9216}

\definecolor{tu130}{rgb}{0.4627, 0.0000, 0.3294}
\definecolor{tu131}{rgb}{0.6118, 0.3020, 0.5333}
\definecolor{tu132}{rgb}{0.7569, 0.5490, 0.6980}
\definecolor{tu133}{rgb}{0.8667, 0.7490, 0.8314}
\definecolor{tu134}{rgb}{0.9216, 0.8510, 0.9020}

\definecolor{con_blue}{HTML}{004488}
\definecolor{con_yellow}{HTML}{DDAA33}
\definecolor{con_red}{HTML}{BB5566}
\definecolor{con_black}{HTML}{000000}
\definecolor{con_gray}{HTML}{AAAAAA}

\definecolor{vib_orange}{HTML}{EE7733}
\definecolor{vib_darkblue}{HTML}{0077BB}
\definecolor{vib_lightblue}{HTML}{33BBEE}
\definecolor{vib_magenta}{HTML}{EE3377}
\definecolor{vib_red}{HTML}{CC3311}
\definecolor{vib_green}{HTML}{009988}
\definecolor{vib_gray}{HTML}{BBBBBB}
\definecolor{vib_black}{HTML}{000000}

\DeclareMathOperator*{\argmin}{arg\,min}

\DeclareMathOperator*{\cls}{class}

\newcommand{\src}[0]{{\mathcal{D}^\mathrm{S}}}
\newcommand{\tgt}[0]{{\mathcal{D}^\mathrm{T}}}

\newcommand{\network}[1]{{\texttt{#1}}}

\newcommand{\dataval}[0]{{\mathcal{D}_\mathrm{val}}}

\newcommand{\datateststar}[0]{{\mathcal{D}_\mathrm{val}}}

\newcommand{\csteststar}[0]{{\mathcal{D}^\mathrm{CS}_\mathrm{val}}}

\newcommand{\bddteststar}[0]{{\mathcal{D}^\mathrm{BDD}_\mathrm{val}}}

\newcommand{\mvteststar}[0]{{\mathcal{D}^\mathrm{MV}_\mathrm{val}}}

\newcommand{\acdcteststar}[0]{{\mathcal{D}^\mathrm{ACDC}_\mathrm{val}}}

\newcommand{\gtavtrain}[0]{{\mathcal{D}^\mathrm{GTA5}_\mathrm{train}}}

\newcommand{\synthiatrain}[0]{{\mathcal{D}^\mathrm{SYN}_\mathrm{train}}}

\pgfplotsset{
table/search path={{./}{../}},
compat = 1.3
}
\graphicspath{{./fig}{../fig}}
\newcommand{\minus}{\scalebox{0.75}[1.0]{$-$}}
\newcommand{\rot}[1]{\rotatebox{90}{#1}}

\newcolumntype{a}{>{\columncolor{tu74}}c}

\usepackage[pagebackref=true,breaklinks=true,letterpaper=true,colorlinks=true,bookmarks=false]{hyperref}
\usepackage[capitalize]{cleveref}
\crefname{section}{Sec.}{Secs.}
\crefname{section}{Section}{Sections}
\crefname{subsection}{Section}{Sections}
\crefname{table}{Table}{Tables}
\crefname{table}{Tab.}{Tabs.}

\def\wacvPaperID{2383} %
\def\confName{WACV}
\def\confYear{2024}

\newcolumntype{g}{>{\columncolor{tu73}}c}
\newcolumntype{h}{>{\columncolor{tu74}}c}
\begin{document}

\title{Generalization by Adaptation: Diffusion-Based Domain Extension \\ for Domain-Generalized Semantic Segmentation}

\author{Joshua Niemeijer$^{1*}$ \qquad Manuel Schwonberg$^{2*}$ \qquad Jan-Aike Termöhlen$^{3*}$\\ \qquad Nico M. Schmidt$^{2}$ \qquad Tim Fingscheidt$^3$\\
$^1$DLR \qquad $^2$CARIAD SE \qquad $^3$Technische Universität Braunschweig\\
{\tt\small joshua.niemeijer@dlr.de \qquad \{manuel.schwonberg, nico.schmidt\}@cariad.technology} \\ {\tt\small \{j.termoehlen, t.fingscheidt\}@tu-bs.de}
}
\maketitle

\begin{abstract}
When models, e.g., for semantic segmentation, are applied to images that are vastly different from training data, the performance will drop significantly. Domain adaptation methods try to overcome this issue, but need samples from the target domain. However, this might not always be feasible for various reasons and therefore domain generalization methods are useful as they do not require any target data.
We present a new diffusion-based domain extension (DIDEX) method and employ a diffusion model to generate a pseudo-target domain with diverse text prompts. In contrast to existing methods, this allows to control the style and content of the generated images and to introduce a high diversity. In a second step, we train a generalizing model by adapting towards this pseudo-target domain. We outperform previous approaches by a large margin across various datasets and architectures without using any real data. For the generalization from GTA5, we improve state-of-the-art mIoU performance by 3.8 \% absolute on average and for SYNTHIA by 11.8 \% absolute, marking a big step for the generalization performance on these benchmarks. Code is available at \small{\url{https://github.com/JNiemeijer/DIDEX}} %
\end{abstract}
\section{Introduction}
\let\thefootnote\relax\footnotetext{* indicates equal contribution}
\label{sec:intro}
\begin{figure}
    \centering
    \includegraphics[width=0.95\columnwidth]{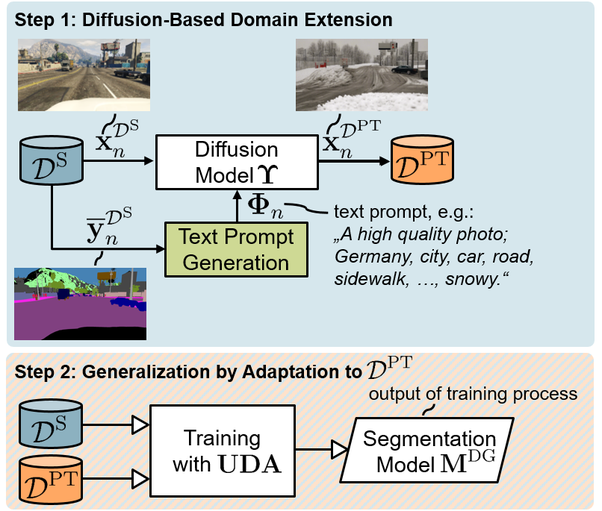}
    \caption{\textbf{Overview} showing a \textbf{block diagram} of our diffusion-based domain extension (DIDEX) method (step 1), followed by the \textbf{generalization-by-adaptation process} to the newly generated pseudo-target domain $\mathcal{D}^\mathrm{PT}$ (step 2).}
    \label{fig:method_overview}
    \vspace{-4.5mm}
\end{figure}
In recent years, the success of deep learning has led to significant advancements in the field of computer vision, e.g., for semantic segmentation. For this task, the usage of synthetic data is particularly interesting as manual data labeling is time- and cost-intensive. Synthetic data can be valuable for training as well as validation, since it allows the simulation of rare and dangerous events.  However, it is still challenging to deploy models trained on synthetic data in real-world settings with varying data distributions due to large domain shifts towards real-world data \cite{tsai2018learning,Niemeijer2023ICCV}.
One approach to overcome this problem is by unsupervised domain adaptation (UDA) \cite{Schwonberg2023Survey}, where unlabeled data from a real target domain is available to adapt to \cite{hoffman2018cycada, yang2020fda,marsden2022contrastive,xie2023sepico,Klingner2020c,Termoehlen2021,Bolte2019a,Niemeijer2022domain,Niemeijer2021Combining}. Some approaches also perform this task source-free~\cite{Klingner2020c} or in a continual manner~\cite{Termoehlen2021,Klingner2020d}. 
Recently, vision transformer models~\cite{Dosovitskiy2021,Xie2022segformer} caused a significant increase in performance and reducing the domain gap with DAFormer \cite{Hoyer2022daformer} being the initial work. 
All these methods rely on access to target domain data to adapt to this particular domain. In practice, this data might not always be available due to various reasons. Data collection can be difficult, because, e.g., adverse weather conditions such as fog, rain, and snow do not constantly occur, and sometimes the target domain cannot be anticipated at all. Consequently, the field of domain generalization (DG) emerged where no data from the target domain is available at all and the task is to generalize from only a single, usually synthetic, source domain to unseen and unknown target domains~\cite{Pan2018,Yue2019,Huang2021,Peng2022semanticaware,Lee2022wildnet,bi2023learning}.
Style transfer methods such as AdaIN~\cite{huang2017arbitrary} are widely used in UDA, especially in combination with other techniques~\cite{hoffman2018cycada, wu2018dcan, wang2020differential}. For domain generalization these methods cannot be used since no target domain guidance is available for the style transfer. For this reason, the majority of DG methods performs style randomization or augmentation to alter the visual appearance of the source domain \cite{Huang2021, peng2021global, kim2021wedge, Yue2019, Lee2022wildnet, sun2023augment}. However, these methods reveal two major issues. First, many of them require additional real data from auxiliary domains~\cite{Lee2022wildnet, kim2021wedge, Huang2021, peng2021global}, which disrupts the idea of domain generalization. Second, the style randomization is difficult to control and often leads to limited diversity, as mostly only textures are changed.\par
In this work, we propose a novel method that tackles the domain generalization problem by diffusion-based domain extension (DIDEX). Diffusion models have demonstrated remarkable capabilities in capturing complex distributions and semantics and generating realistic samples with high quality \cite{Rombach2022Stable, ho2020denoising, song2020denoising, dhariwal2021diffusion}. 
Our method allows not only to generate realistic pseudo-target images from synthetic images but also to alter important parameters such as location, time, weather conditions and semantic content via text prompts. We use these capabilities to generate a realistic looking pseudo-target domain as shown in Figure~\ref{fig:method_overview}.
Diffusion models have severe limitations in the semantic consistency of the generated output \cite{zhang2023adding} w.r.t.\ the input image. To address these limitations, we draw inspiration from the field of \textit{unsupervised} domain adaptation (UDA) and define the diffusion data as the target domain for our adaptation process. 
The key contributions of this paper are as follows: 
\begin{itemize}
\itemsep0.2em
    \item We propose a new method leveraging diffusion models for domain transfer, enhancing our model's ability to generalize across domains without accessing real data.
    \item By utilizing UDA techniques, we overcome the semantic inconsistency limitation of diffusion models, ensuring improved domain generalization performance.
    \item We report results on common benchmarks, where our method outperforms the state-of-the-art domain generalization methods by a large margin.
\end{itemize} 
\section{Related Work}
\label{sec:related-work}
 In this section, we will give a brief overview on the related methods for unsupervised domain adaptation, domain generalization, and diffusion models. \par
\subsection{Unsupervised Domain Adaptation (UDA)}
\begin{figure*}[t]
    \centering
    \includegraphics[width=1.4\columnwidth]{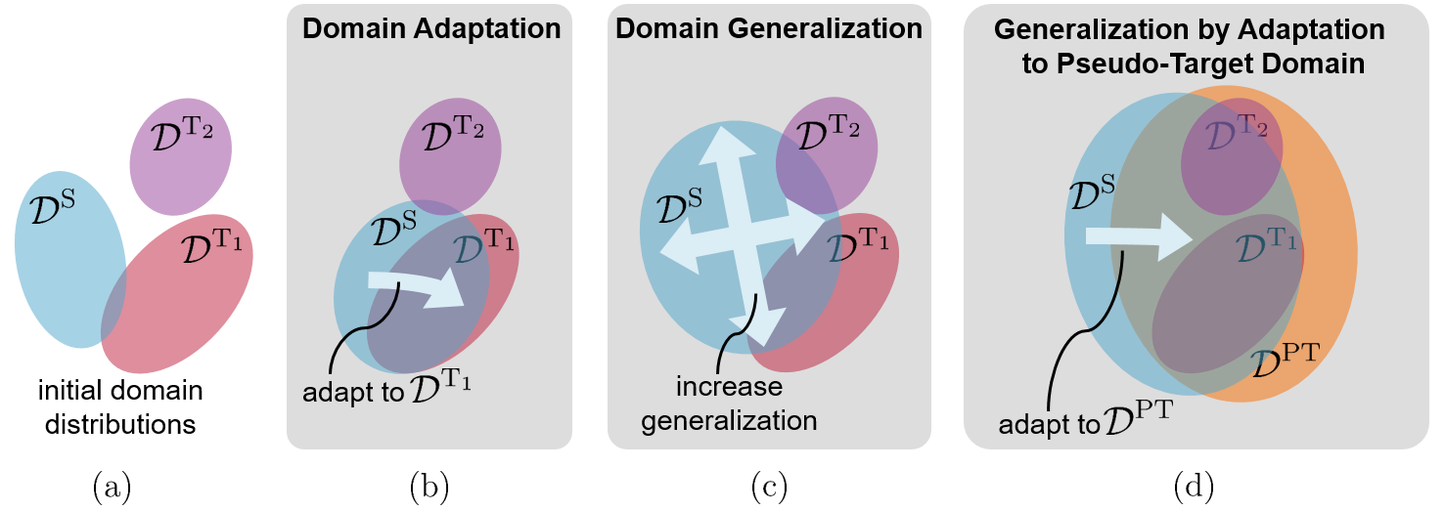}
    \caption{\textbf{Simplified representation of different domain distributions} and the effects of domain adaptation and generalization methods. Shown in (a) are the initial distributions for a source domain $\src$ and two target domains ${\mathcal{D}^\mathrm{T_1}}$ and ${\mathcal{D}^\mathrm{T_2}}$. In (b), the effect of domain adaptation from $\src$ to a specific target domain ${\mathcal{D}^\mathrm{T_1}}$ is shown. In (c), the effect of commonly used domain generalization methods is shown that broaden the source distribution, e.g., via data augmentation, but in a less directed manner. In (d), we show the effect of our proposed method, were we perform a domain extension and generate a pseudo-target domain ${\mathcal{D}^\mathrm{PT}}$ that also covers the distributions of multiple target domains (${\mathcal{D}^\mathrm{T_1}}$ and ${\mathcal{D}^\mathrm{T_2}}$). For the purpose of clarity, only $K\!=\!2$ target domains are shown here. In the remainder of this paper, by default, we employ $K\!=\!4$ target domains for the evaluation. 
    }
    \label{fig:distributions}
\end{figure*}
UDA methods are clustered into input, feature, output and hybrid adaptation approaches \cite{Schwonberg2023Survey}. Our method has similarities to the idea of UDA input space adaptation where the aim is to style transfer the visual appearance between the source and target domain. Often a CycleGAN \cite{zhu2017unpaired} is utilized for a semantically consistent style transfer, usually in conjunction with other methods \cite{dong2021and, hoffman2018cycada, gong2019dlow, xu2022self, li2019bidirectional}. In the feature space, contrastive learning methods such as SePiCo \cite{xie2023sepico}, UCDA \cite{zhang2022unsupervised}, ProDA \cite{zhang2021prototypical}, CPSL \cite{li2022class} and CONFETI \cite{li2023contrast} obtained significant improvements. DAFormer \cite{Hoyer2022daformer}, HRDA \cite{hoyer2022hrda} and MIC \cite{hoyer2023mic} are transformer-based approaches improving the state-of-the-art performance significantly by using a slightly modified \network{SegFormer} \cite{Xie2022segformer} architecture and employing methods such as rare class sampling, feature distance, teacher-student training, self-training, multi-resolution training and masked image consistency. Benigmim et al.\ ~\cite{Benigmim2023_personalizeddiff} proposed a data augmentation technique using a finetuned text-to-image diffusion model for one-shot unsupervised domain adaptation.
\subsection{Domain Generalization (DG)} Domain generalization methods for semantic segmentation try to overcome the domain gap by either constraining learned feature distributions or randomizing distributions by augmentation or extension during training. \par
Pan \etal proposed IBN-Net~\cite{Pan2018} as one of the first DG approaches and employed instance-batch normalization (IBN), which is more robust w.r.t. appearance changes. Further approaches utilize instance whitening to decorrelate features from different layer channels and thereby removing style-related information. Pan \etal \cite{pan2019switchable} proposed an approach which can switch between different whitening and normalization techniques depending on the task. RobustNet ~\cite{Choi2021}, DIRL~\cite{xu2022dirl}, and SAN+SAW~\cite{Peng2022semanticaware} all proposed improvements such as a better guidance of the whitening process by a so-called sensitivity-aware prior module \cite{xu2022dirl} or semantic-aware whitening and normalization \cite{Peng2022semanticaware}.\par
Domain randomization can be separated into approaches that require additional real-world data (auxiliary domains) and those which do not need it.
Yue \etal~\cite{Yue2019} proposed one of the first domain randomization methods combining domain randomization and pyramid consistency (DRPC), enforcing semantic feature consistency between differently stylized images with transferred textures from ImageNet. Similarly, Huang \etal ~\cite{Huang2021} with FSDR employ style randomization in the frequency space, Peng \etal~\cite{peng2021global} use painting for their style randomization and WildNet by Lee \etal~\cite{Lee2022wildnet} applies style randomization in conjunction with contrastive and consistency learning. Kim \etal~\cite{kim2021wedge} utilize style randomization with internet sampled images and self-training similar to UDA approaches. In contrast to these methods, our proposed method can be seen as a more structured distribution randomization without the need for real-world auxiliary domains. \par
Only few works perform style randomization without additional real data. AugFormer by Schwonberg \etal \cite{schwonberg2023augmentation} showed that simple augmentations can significantly increase domain generalization. Similarly, Sun \etal \cite{sun2023augment} proposed a strong style randomization in the CIELAB color space. Zhong \etal \cite{zhong2022adversarial} proposed to alter the channel-wise mean and standard deviation in an adversarial manner to generate stylized images which are hard to segment for the model. SHADE by Zhao \etal~\cite{zhao2022style} introduces a style hallucination module which is based on farthest point sampling and generates new diverse samples by a linear combination of the basic styles. Bi \etal \cite{bi2023learning} proposed CMFormer which is the first DG approach tailored towards vision transformers. It fuses low and original resolution in a so-called content enhanced mask attention. In contrast, our approach is independent from the architecture.
Gong \etal \cite{gong2023prompting} proposed PromptFormer as the first DG approach utilizing diffusion models. However, their method fundamentally differs from ours since they employ a diffusion backbone for domain-invariant pre-training, where scene and category text prompts help the model to disentangle domain-variant and invariant information. They also employ a consistency loss to enforce same predictions under different input prompts. 
\subsection{Diffusion Models} 
Diffusion models for image synthesis recently emerged and outperform the long established standard using GANs \cite{dhariwal2021diffusion}. Several improvements were proposed to accelerate the image generation, namely denoising diffusion probabilistic models (DDPMs) \cite{ho2020denoising} and denoising diffusion implicit models (DDIMs) \cite{song2020denoising}. Rombach \etal proposed latent diffusion models (LDMs) sometimes referred to as stable diffusion models which reduce the training and inference time and enable text-to-image and image-to-image generation. Based on this, several extensions and improvements were proposed such as stable diffusion XL (SDXL) \cite{podell2023sdxl} or \network{ControlNet} \cite{zhang2023adding}, which offers the possibility to constrain the generation process with different additional inputs such as depth or semantics.

\section{Proposed Method}
\label{sec:proposed-method}
In this section, we will describe the approach on the text prompt generation of our new diffusion-based domain extension (DIDEX) method. Subsequently, we will describe the generalization-by-adaptation approach.

\subsection{Diffusion-Based Domain Extension}
\label{sec:didex}
The general approach and text prompt generation will be introduced in the following in detail. 
\subsubsection{General Approach}
We define an input image as $\mathbf{x}_n \in \mathbb{G}^{H \times W \times C}$, with $\mathbb{G}$ denoting the set of integer color intensity values, $H$ and $W$ image height and width in pixels, and $C\!=\!3$ the number of color channels, respectively. Images in a dataset with $N$ images are indexed $n \in \{1,2,...,N\}$. A semantic segmentation network $\VEC{M}$ maps $\mathbf{x}_n$ to an output map $\mathbf{y}_n = (y_{n,i,s})\in \mathbb{I}^{H \times W \times S}$ with posterior probabilities $y_{n,i,s} = P(s|i,\mathbf{x}_n)$ for each class $s \in \mathcal{S}$ at pixel index $i \in \mathcal{I} = \{1,2,...,H\cdot W\}$. Here, $\mathcal{S} = \{1,2,...,S\}$ denotes the set of $S$ classes and $\mathbb{I}=[0,1]$.
Superscripts ``$\mathrm{S}$'' and ``$\mathrm{T}$'' on $\mathbf{x}_n$ and $\mathbf{y}_n$ denote the domain from which the variables stem, with, e.g., ${\src}$ being the source domain and ${\tgt}$ being the target domain. 
With domain generalization, the aim is to generalize to multiple unseen domains so the target domains $\mathcal{D}^{\mathrm{T}_k}$ are indexed with $k \in \mathcal{K} = \{1,2,...,K\}$, where $K$ is the total amount of target domains that the method will be evaluated on. 
The crucial advantage of diffusion models is the prompt-based control of the generated image. That enables us to define the content and style of the generated images and create a pseudo-target domain $\mathcal{D}^{\mathrm{PT}}$. Two different strategies for the generation of pseudo-target domains are possible. First, a specialized pseudo-target domain could be created, if some information regarding the target domains are available, such as location, weather, etc. In our approach, however, we aim for a domain generalization without any assumptions about the target domains and generate $\mathcal{D}^{\mathrm{PT}}$ with high diversity to cover a wide range of possible target domain distributions as shown in Figure \ref{fig:distributions} (d).
We formulate the objective to generate a pseudo-target domain, which is the union of all relevant target domains as
\begin{equation}
    \mathcal{D}^{\mathrm{PT}} \approx \bigcup_{k\in\mathcal{K}} \mathcal{D}^{\mathrm{T}_k} \, ,
\end{equation}
where $\mathcal{D}^{\mathrm{T}_k}$ denotes the relevant target domains. 
In our domain generalization setting we only have a single source domain distribution available to obtain $\mathcal{D}^{\mathrm{PT}}$. We use the diffusion model to extend the initial source distribution into the more diverse and wider pseudo-target distribution. A single image of $\mathcal{D}^{\mathrm{PT}}$ can be obtained by:
\begin{equation}
    \mathbf{x}_n^{\mathrm{PT}} = \VECG{\Upsilon}(\mathbf{x}_n, \VECG{\Phi}_n)
\end{equation}
where $\VECG{\Upsilon}$ denotes the diffusion model and $\VECG{\Phi}_n$ the text prompt. The generated pseudo-target domain provides the crucial advantage that we have a target domain available that can be used for adapting the segmentation network as shown in Figure \ref{fig:distributions} (d). Compared to other DG approaches, this is a novel and unique feature of our method and enables us to obtain generalization by adaptation.\par
\textbf{Semantic consistency:} We observed that semantic or structural consistency of the image-to-image generation is often not given for complex urban street scenes, i.e., pedestrians or cars can disappear and the semantic content of the pixels is changed (see Figure \ref{fig:data-vis-main}). Directly using the synthetic labels for the pseudo-target domain is therefore not possible. However, we assume that a certain structural and semantic consistency is also beneficial for the adaptation in step two and for this reason we employ depth-guided \network{Stable Diffusion 2.0} \cite{Rombach2022Stable} and \network{ControlNet} \cite{zhang2023adding}, which offer a higher structural and semantic consistency. Both extend the image generation to
\begin{equation}
    \mathbf{x}_n^{\mathrm{PT}} = \VECG{\Upsilon}(\mathbf{x}_n, \VEC{C}(\mathbf{x}_n), \VECG{\Phi}_n) \, ,
\end{equation}
where $\VEC{C}$ is a constraint based on the image $\mathbf{x}_n$, e.g., by canny edge extraction, depth or  semantic prediction. %
\subsubsection{Text Prompt Generation}
\begin{figure}
    \centering
    \includegraphics[width=\columnwidth]{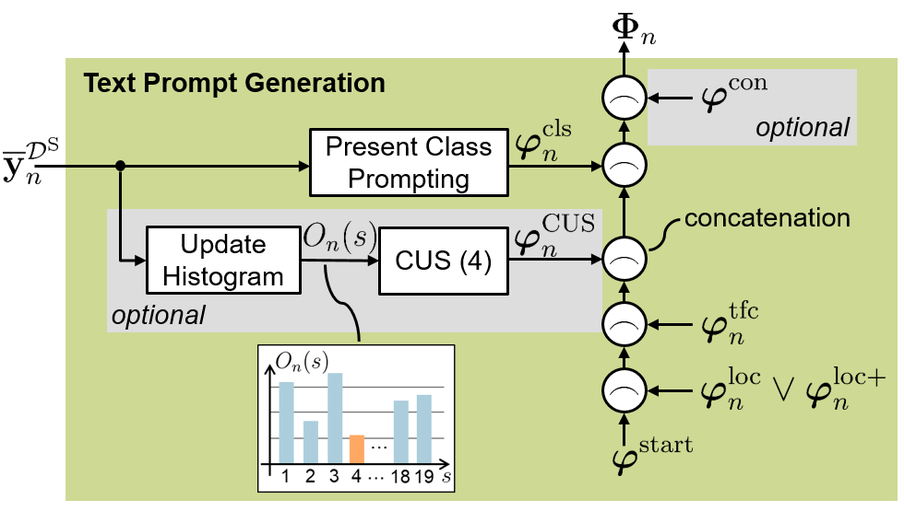}
    \caption{Block diagram of the \textbf{text prompt generation} from Figure~\ref{fig:method_overview} providing $\VECG{\Phi}_n$. The operation in the circles denotes a concatenation. Concatenation of $\VECG{\varphi}^{\mathrm{CUS}}$ and $\VECG{\varphi}^{\mathrm{con}}$ is optional. Class-uniform sampling (CUS) is described in (\ref{eq:cus}). The rarest class in the example histogram $O_n(s)$ is marked in orange.}
    \label{fig:method}
\end{figure}
Text prompts are crucial for the class distribution of the pseudo-target domain, since the prompts determine the style and content of the generated images.  
To the best of our knowledge, there are no existing strategies how to design text prompts for diffusion models in the context of complex urban street scenes. For this reason, we developed a new, systematic and modular text prompt generation that is visualized in Figure~\ref{fig:method}. Our text prompt strategy aims at varying the image content and style at different levels, introducing as much diversity as possible into the pseudo-target domain.
To design the building blocks of the prompts, we considered major causes of real-to-real domain shifts as a guidance.\par
The start of each prompt is $\VECG{\varphi}_n^{\mathrm{start}} = \textit{``A high quality photo;''}$
We observed that particularly emphasizing photorealism with prompts such as "real" or "photorealistic" did not improve the realism of the output.\\
The location shift between different countries or continents is important in automated driving so we include a set of different countries and regions to include their style and content in $D^{PT}$.  We concatenate a geographic location to the start prompt by choosing either a location $\VECG{\varphi}_n^{\mathrm{loc}} \in \{\textit{``Europe''},\textit{``Germany''}\}$
or a location from an extended location set
$\VECG{\varphi}_n^{\mathrm{loc+}} \in \{\textit{``Europe''},\textit{``Germany''},\textit{``USA''},\textit{``China''},\textit{``India''}\}.$ We are aware that this set of locations is not globally comprehensive and may have a significant impact on the domain generalization performance. We leave an extensive study with more regions of the world for future work.
Subsequently we concatenate a traffic location 
$\VECG{\varphi}_n^{\mathrm{tfc}} \in \{\textit{``\_''} \text{(blank)},\textit{``highway''},\textit{``city''}\}.$ The blank denotes that no traffic location is provided.\par
\textbf{Present class prompting:}
We utilize the GTA5 segmentation label map $\overline{\mathbf{y}}_n^{\mathcal{D}^\mathrm{S}}$ to identify and to concatenate all classes that are present in the current image. This serves as an additional guidance on which classes $\VECG{\varphi}_n^{\mathrm{cls}} \subseteq \mathcal{S}$ the diffusion model should include in the output.\par
\textbf{Class-uniform sampling (CUS):} 
Optionally, we introduce a class-uniform sampling (CUS) to mitigate the unbalanced class distribution by tracking the number of times a class occurred in the synthetic source images and concatenating the least often seen class to the prompt. For each image with index $n$, we utilize the segmentation labels from the respective source image $\overline{\mathbf{y}}_n^{\mathcal{D}^\mathrm{S}}$ to update a histogram with the class occurrences $O_n(s)$. We then concatenate the rarest class name to the prompt: 
\begin{equation}
    \label{eq:cus} \VECG{\varphi}_n^{\mathrm{CUS}} = \cls(\argmin_{s\in\mathcal{S}}(O_n(s))).
\end{equation}
After concatenating the rarest class to our prompt, we subsequently increase the counter for this class. 
This leads to a more uniform class distribution in our domain extension.

\textbf{Image conditions:}
Adverse weather or visibility conditions are a challenge for autonomous vehicles and we therefore also represent these shifts in our pseudo-target domain by adding one of the following words to the prompt:
\begin{align*}\VECG{\varphi}_n^{\mathrm{con}} \in \{&\textit{``rain''},\textit{``fog/mist''},\textit{``snowy''},\textit{``sunny''},\\
&\textit{``overcast''},\textit{``stormy''},\textit{``overexposure''},\\&\textit{``underexposure''},\textit{``evening''},\textit{``morning''},\\
&\textit{``night/darkness''},\textit{``backlighting''},\\
&\textit{``artificial lighting''},\textit{``harsh light''},\\
&\textit{``dappled light''},\textit{``sun flare''},\textit{``hazy/haze''},\\
&\textit{``spring''},\textit{``autuum''},\textit{``winter''},\textit{``summer''}\}\end{align*}
By concatenating all the different building blocks, we obtain our text prompt according to Figure~\ref{fig:method} as: 
\begin{multline}
    \VECG{\Phi}_n= \VECG{\varphi}^{\mathrm{start}} \frown (\VECG{\varphi}_n^{\mathrm{loc}}\lor\VECG{\varphi}_n^{\mathrm{loc+}}) \frown \VECG{\varphi}^{\mathrm{tfc}} \\ \frown \VECG{\varphi}^{\mathrm{CUS}}\frown
    \VECG{\varphi}^{\mathrm{cls}}\frown 
    \VECG{\varphi}^{\mathrm{con}}\, ,
\end{multline}
with $\frown$ symbolizing concatenation and $\VECG{\varphi}^{\mathrm{CUS}}$ and $\VECG{\varphi}^{\mathrm{con}}$ being optional. We denote the base prompt without the optional strings as $\VECG{\Phi}_n^{\mathrm{base}}$.
A sample prompt can become, e.g., \textit{``A high quality photo; Europe, highway, road, car, building, vegetation, winter''}. 
If there are multiple options within one of the building blocks, we select the prompt randomly. We generate $\mathcal{D}^{\mathrm{PT}}$ before the adaptation step.

\subsection{Generalization by Adaptation}
Even with strong semantic guidance by depth or the semantic ground truth, the stable diffusion models generate partially highly inconsistent outputs.
However, the area of unsupervised domain adaptation provides powerful methods which are capable to handle datasets without labels effectively. We utilize these methods to adapt the segmentation model towards the pseudo-target domain that was generated in a 1\textsuperscript{st} step (Section~\ref{sec:didex}). For UDA methods, we usually obtain an adapted segmentation model by $ \VEC{UDA}(\src, \tgt)\!\rightarrow\!\VEC{M}^\mathrm{UDA}$, where $\tgt$ is usually a real data target domain without labels and $\VEC{M}$ our domain-adapted model. Our method changes this setting slightly but notably to $ \VEC{UDA}(\src, \mathcal{D}^{\mathrm{PT}}) \!\rightarrow\!\VEC{M}^\mathrm{DG}$, where model $\VEC{M}$ is adapted to our pseudo-target domain $\mathcal{D}^{\mathrm{PT}}$ and therefore also generalizes without accessing any real data. In our case, UDA strategies create this supervision signal by aligning the distributions of the source $\mathcal{D}^\mathrm{S}$ and pseudo-target domain $\mathcal{D}^{\mathrm{PT}}$.
\section{Experimental Setup}
\label{sec:experimental-setup}
In the following, we introduce the employed datasets, metrics, and diffusion and segmentation network architectures. Afterwards, we introduce the utilized UDA methods.
\subsection{Datasets and Metrics}
\par \textbf{Datasets:}
We follow the common domain generalization standard setting~\cite{gong2023prompting, Lee2022wildnet, sun2023augment, Peng2022semanticaware, peng2021global} and employ the two synthetic datasets SYNTHIA (SYN)\cite{Ros2016} and GTA5 \cite{Richter2016} as our source domains. The datasets comprise 9400 and 24966 images, respectively, which we denote as $\synthiatrain$ and $\gtavtrain$. \\
To evaluate the domain generalization capabilities, we use Cityscapes (CS) \cite{Cordts2016}, BDD100k (BDD) \cite{Yu2018b}, Mapillary Vistas (MV)\cite{Neuhold2017} as our real-world target domains with 500, 1000 and 2000 validation images, respectively. The training sets of these datasets remain unused in our generalization method. We evaluate all experiments on the validation sets $\dataval$ of our target domains and compute the domain generalization (DG) mean over these sets, as is common practice in domain generalization~\cite{Yue2019, Huang2021,Peng2022semanticaware}. As an addition to this domain generalization benchmark, we also include ACDC \cite{Sakaridis2021acdc} with 406 validation images in some of our experiments, which contains adverse weather conditions. It is excluded from the DG mean, because most other approaches do not report ACDC performance.
\par \textbf{Metrics:} For evaluation we employ the mean intersection over union (mIoU) of $S\!=\!19$ segmentation classes~\cite{Cordts2016,Richter2016,Sakaridis2021acdc} for GTA5 trained models and $S\!=\!16$ classes for models trained on SYNTHIA, as it is common practice~\cite{Klingner2020c}.
\begin{table}[t!]
  \centering
  \renewcommand{\arraystretch}{.8}
  \setlength{\tabcolsep}{.05em}
  \caption{\textbf{Domain generalization performance} (mIoU (\%)) of several methods employing two different encoder networks. \textbf{Training} was performed on the synthetic \textbf{GTA5} ($\src\!=\!\gtavtrain$) dataset. \textbf{Evaluation} is performed on various real-world \textbf{validation sets} ($\tgt\!=\!\datateststar$). Prior work results are either cited from~\cite{Lee2022wildnet} (marked with $^\circ$) or from the respective paper (marked with *). 
  }
  
\extrarowheight=\aboverulesep
    \addtolength{\extrarowheight}{\belowrulesep}
    \aboverulesep=0pt
    \belowrulesep=0pt
    \begin{tabular}{@{}clchhhc@{}}
        \toprule
        \multirow{1}{*}{\rot{\makecell[c]{\textbf{Enc.\hphantom{0}}}}} & \textbf{DG Method} &\multirow{1}{*}{\makecell{\textbf{No} \\[-0.15em]\textbf{real} \\[-0.15em] \textbf{data}}}&\multicolumn{4}{c}{\textbf{mIoU (\%) on}} \\
        \cmidrule{4-7}
       &&     & $\csteststar$ & $\bddteststar$ & $\mvteststar$ & \raisebox{-4.0pt}{\shortstack{\textbf{DG} \\\textbf{mean}}}  \rule[-1.2ex]{0mm}{3.65ex} \\

        \midrule
        \parbox[t]{2mm}{\multirow{13}{*}{\rotatebox[origin=c]{90}{\small{\network{ResNet}}}}}
        &Baseline  & \ding{51} & 36.1 & 36.6 & 43.8& 38.8\\
        &IBN-Net$^\circ$~\cite{Pan2018}  & \ding{51} & 37.7 & 36.7 & 36.8 & 37.1\\
        &RobustNet$^\circ$~\cite{Choi2021}& \ding{51} & 37.3 & 38.7 & 38.1 &38.0\\
        &DRPC*~\cite{Yue2019}& \ding{55} & 42.5& 38.7& 38.1&39.8\\
        &SW*~\cite{pan2019switchable}& \ding{51} & 36.1 &36.6&32.6&35.1 \\
        &FSDR*~\cite{Huang2021}& \ding{55} & 44.8 &41.2&43.4&43.1 \\
        &SAN+SAW*~\cite{Peng2022semanticaware}& \ding{51} & 45.3 &41.2& 40.8 &42.4\\
       &WEDGE* \cite{kim2021wedge}&\ding{55}& 45.2&41.1&48.1&44.8 \\
       &GTR* \cite{peng2021global}&\ding{55}&43.7 &39.6&39.1&40.8 \\
        &SHADE* \cite{zhao2022style}&\ding{51}&46.7 &43.7&45.5&45.3 \\
        &WildNet$^\circ$~\cite{Lee2022wildnet}  & \ding{55} & {45.8} & {41.7} & {47.1} &44.9\\
        &RICA*~\cite{sun2023augment}  & \ding{51} & {48.0} & \textbf{45.2} & {46.3} & {46.5}\\
         &DIDEX with MIC~\cite{hoyer2023mic} & \ding{51} & {\textbf{52.4}} & {40.9} & {\textbf{49.2}} &\textbf{47.5}\\
       
        \midrule
        \parbox[t]{2mm}{\multirow{7}{*}{\rotatebox[origin=c]{90}{\small{\network{Transformer}}}}} &Baseline &\ding{51} & 46.6&45.6& 50.1   &47.4\\
        &ReVT* \cite{Termoehlen2023}   &\ding{51} & 50.0&48.0& 52.8   &50.3\\
        &DAFormer* \cite{Hoyer2022daformer}   &\ding{51} & 52.7&47.9& 54.7  &51.7\\
        &HRDA* \cite{hoyer2023domain}                 &\ding{51} & 57.4& 49.1& 61.2&55.9\\
        &CMFormer* \cite{bi2023learning}   & \ding{51}& 55.3& 49.9& 60.1&55.1\\
        &PromptFormer* \cite{gong2023prompting}  & \ding{51}& 52.0& -& -&- \\

        &DIDEX with MIC~\cite{hoyer2023mic} & \ding{51}& \textbf{62.0} & \textbf{54.3} & \textbf{63.0} &\textbf{59.7}\\

        \bottomrule
    \end{tabular}

  \label{tab:sota_GTA} 
   \vspace{-4.5mm}
\end{table}
\begin{table}[t!]
  \centering
  \renewcommand{\arraystretch}{.8}
  \setlength{\tabcolsep}{.05em}
  \caption{\textbf{Domain generalization performance} (mIoU (\%)) of several methods employing two different encoder networks. \textbf{Training} was performed on the synthetic \textbf{SYNTHIA} ($\src\!=\!\synthiatrain$) dataset. \textbf{Evaluation} is performed on real-world \textbf{validation sets} ($\tgt\!=\!\datateststar$).  Prior work results are either cited from~\cite{Lee2022wildnet} (marked with $^\circ$) or from the respective paper (marked with *).}
  
\extrarowheight=\aboverulesep
    \addtolength{\extrarowheight}{\belowrulesep}
    \aboverulesep=0pt
    \belowrulesep=0pt
    \begin{tabular}{@{}clchhhc@{}}
        \toprule
        \multirow{1}{*}{\rot{\makecell[c]{\textbf{Enc.\hphantom{0}}}}} & \textbf{DG Method} &\multirow{1}{*}{\makecell{\textbf{No} \\[-0.15em]\textbf{real} \\[-0.15em] \textbf{data}}}&\multicolumn{4}{c}{\textbf{mIoU (\%) on}} \\
        \cmidrule{4-7}
       &&     & $\csteststar$ & $\bddteststar$ & $\mvteststar$ & \raisebox{-4.0pt}{\shortstack{\textbf{DG} \\\textbf{mean}}}  \rule[-1.2ex]{0mm}{3.65ex} \\
        \midrule
        \parbox[t]{2mm}{\multirow{10}{*}{\rotatebox[origin=c]{90}{\small{\network{ResNet}}}}} 
        &Baseline  & \ding{51} & 34.3 & 27.8 & 38.0 & 33.4\\
       
        &IBN-Net$^\circ$~\cite{Pan2018}  & \ding{51} & 34.2 & 32.6 & 36.2 & 34.3\\
        &DRPC*~\cite{Yue2019}& \ding{55} & 37.6& 34.4& 34.1&35.4\\
        &SW*~\cite{pan2019switchable}& \ding{51} & 36.1 &36.6&32.6&35.1 \\
        &FSDR*~\cite{Huang2021}& \ding{55} & 40.8 &37.4&39.6&39.3 \\
        &SAN+SAW*~\cite{Peng2022semanticaware}& \ding{51} & 40.9 &36.0& 37.3 &38.1\\
        &WEDGE* \cite{kim2021wedge}&\ding{55}& 40.9&38.1&43.1&40.7 \\
        &GTR* \cite{peng2021global}&\ding{55}&39.7 &35.3&36.4&37.1 \\
        &RICA*~\cite{sun2023augment}  & \ding{51} & {45.0} & {36.3} & {41.6} & {41.0}\\
        &DIDEX with MIC~\cite{hoyer2023mic}& \ding{51} & {\textbf{53.1}} & \textbf{41.8} & {\textbf{50.3}} &\textbf{48.4}\\
        \midrule
        \parbox[t]{2mm}{\multirow{5}{*}{\rotatebox[origin=c]{90}{\small{\network{Transformer}}}}} &Baseline    &\ding{51} & 41.4&36.2& 42.4  &40.0\\
        &ReVT* \cite{Termoehlen2023}   &\ding{51} & 46.3&40.3& 44.8   &43.8\\
        &CMFormer* \cite{bi2023learning}   & \ding{51}& 44.6& 33.4& 43.3&40.4\\
        &PromptFormer* \cite{gong2023prompting}  & \ding{51}& 49.3& -& -&- \\
        &DIDEX with MIC~\cite{hoyer2023mic}& \ding{51}& \textbf{59.8} & \textbf{47.4} & \textbf{59.5} &\textbf{55.6}\\  
        \bottomrule
    \end{tabular}

  \label{tab:sota_synthia} 
   \vspace{-4.5mm}
\end{table}
\subsection{Network Architectures}
\textbf{Diffusion models:}
For the data generation process, we employ \network{Stable Diffusion 2.0} (\network{SD2.0}) \cite{Rombach2022Stable} as our standard diffusion model. We also compare against \network{ControlNet} \cite{zhang2023adding}, which provides several options to constrain the output generation in our ablation  studies. 
To create data, we utilize the image-to-image prompting strategy described in Section~\ref{sec:proposed-method}. 
The ablation studies w.r.t. the data generation process that are described in Section~\ref{sec:experiments:propting} analyze the different prompting elements.\par
\par \textbf{Segmentation models:}
As the network architectures we selected the common standards in the area of domain generalization. The \network{DeepLabV2}~\cite{chen2017deeplab} architecture with a \network{ResNet-101}  is used to evaluate the generalization on a CNN-based architecture and for the recently emerged vision transformer we use the \network{DAFormer} \cite{Hoyer2022daformer} network.  
\subsection{Employed UDA Methods}
We utilize five different state-of-the-art domain adaptation approaches for our generalization by adaptation step, namely DACS \cite{tranheden2021dacs}, DAFormer \cite{Hoyer2022daformer}, HRDA \cite{hoyer2022hrda}, MIC \cite{hoyer2023mic} and SePiCo \cite{xie2023sepico}.
As described in Section \ref{sec:proposed-method}, we utilize these UDA methods to adapt from the synthetic source domain to our pseudo-target domain.
Previous experiments have shown that the adaptation of UDA methods to other domain shifts can diminish the performance \cite{Sakaridis2021acdc}. In this context, \textit{it has to be noted that we do not finetune any hyperparameters of the employed UDA methods.} All the methods are employed as provided by the respective authors. 
\begin{figure*}
    \centering
    \includegraphics[width=0.9\linewidth]{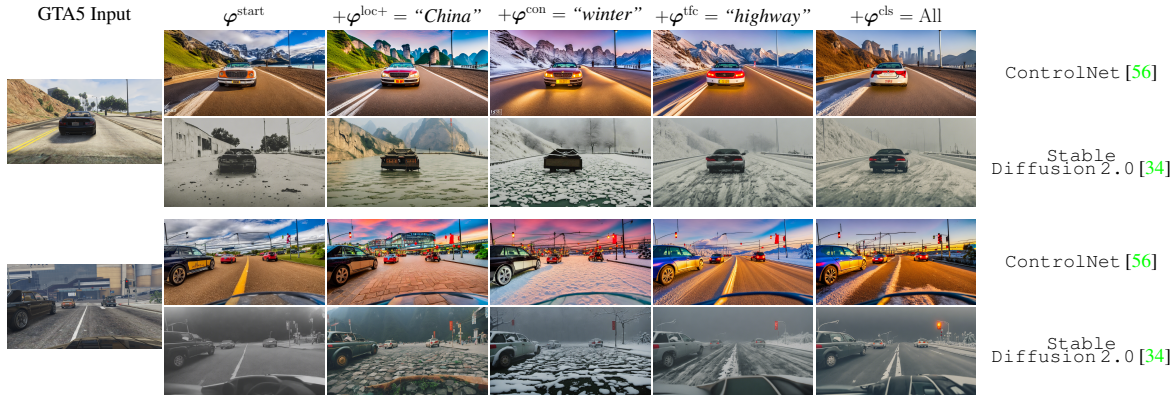}
    \caption{\textbf{Visualization of the prompt ablation study} showing the impact of the prompt on the images of $\mathcal{D}^{\mathrm{PT}}$ for \network{ControlNet} \cite{zhang2023adding} and \network{SD 2.0} \cite{Rombach2022Stable}. Each prompt has a corresponding impact on the content of the generated images. However, semantic inconsistencies can be seen such as trees turning into mountains or buildings which disappear. More examples are provided in the supplement.}
    \label{fig:data-vis-main}
\end{figure*}
\section{Evaluation and Discussion}
\label{sec:experiments:propting}
\begin{table*}
  \centering
  \renewcommand{\arraystretch}{.725}
  \setlength{\tabcolsep}{.35em}
  \caption{\textbf{Influence of the different text prompt building blocks} on the generalization performance (mIoU(\%)). \textbf{Training} was performed on the synthetic \textbf{GTA5} ($\src\!=\!\gtavtrain$) dataset. \textbf{Evaluation} is performed on various real-world \textbf{validation sets} ($\tgt\!=\!\datateststar$). The adaptation was performed with the HRDA method~\cite{hoyer2022hrda} and a \network{Transformer}-based encoder.}
  
    \extrarowheight=\aboverulesep
    \addtolength{\extrarowheight}{\belowrulesep}
    \aboverulesep=0pt
    \belowrulesep=0pt
        \begin{tabular}{cccchhhcc}
        \toprule
        \multirow{1}{*}{\makecell{\textbf{Base Prompt} \\($\VECG{\Phi}^{\mathrm{base}}$)}}& \multirow{1}{*}{\makecell{\textbf{Add. Location}\\ ($\VECG{\varphi}^{\mathrm{loc}}\to\VECG{\varphi}^{\mathrm{loc+}}$)}} & \multirow{1}{*}{\makecell{\textbf{+ Conditions}\\ ($+\VECG{\varphi}^{\mathrm{con}}$)}}& \multirow{1}{*}{\makecell{\textbf{+ CUS}\\ ($+\VECG{\varphi}^{\mathrm{CUS}}$)} }& \multicolumn{5}{c}{\textbf{mIoU (\%) on}} \\
        \cmidrule{5-9} 
       & &&&$\csteststar$ & $\bddteststar$ & $\mvteststar$ & $\acdcteststar$ & \raisebox{-4.0pt}{\shortstack{\textbf{DG} \\\textbf{mean}}}  \rule[-1.2ex]{0mm}{3.65ex}\\
        \midrule
        \checkmark & \minus     & \minus     & \minus     & 58.5 & 52.2 & 62.9 & 46.9 & 57.9 \\
        \checkmark & \checkmark & \minus     & \minus     & 58.7 & 52.5 & 63.4 & 46.7 & 58.2 \\
        \checkmark & \minus     & \checkmark & \minus     & 59.4 & {52.7} & 62.7 & 46.8 & 58.3 \\
        \checkmark & \minus     & \minus     & \checkmark & \textbf{61.2} & 52.5 & \textbf{63.7} & \textbf{48.8} & \textbf{59.1} \\ %
        \checkmark &   \checkmark     & \minus     & \checkmark & 58.6 & 51.8 & 62.8 & 45.2 & 57.7 \\
        \checkmark & \minus     & \checkmark     & \checkmark & 60.1 & \textbf{53.7} & 63.5 & 46.6 & \textbf{59.1} \\
        \checkmark & \checkmark & \checkmark & \checkmark & 58.8 &{52.7} & 63.2 & 47.4 & 58.3 \\
        \bottomrule
        \end{tabular}

  \label{tab:prompt_abl} 
   \vspace{-4.5mm}
\end{table*}
In this section, we will first compare our approach to state-of-the-art domain generalization methods. Afterwards, we investigate the impact of different prompting strategies, UDA approaches, semantic consistency constraints, and the quantity of generated images.
\subsection{Comparison with State of the Art}
We compare the generalization performance of our method to state-of-the-art methods in Table \ref{tab:sota_GTA} ($\src\!=\!\gtavtrain$) and Table~\ref{tab:sota_synthia} ($\src\!=\!\synthiatrain$).
For GTA5-trained models, for both the \network{ResNet}-based and the \network{Transformer}-based backbones, we clearly achieve a new state-of-the-art (SOTA) performance for the DG mean. 
For the \network{ResNet}-based models, we improve significantly on Cityscapes and Mapillary. 
For SYNTHIA-trained models (Table~\ref{tab:sota_synthia}) our method provides an even larger improvement. With a \network{ResNet}-based backbone we outperform other approaches by 7.4\% absolute for the DG mean and by 11.8\% abs.\ with a transformer backbone. On Mapillary Vistas, our method provides 14.7\% abs.\ mIoU improvement compared with the best performing prior method \cite{Termoehlen2023}. With 59.8\% mIoU on Cityscapes, DIDEX performs competitive with the UDA method DAFormer~\cite{Hoyer2022daformer} with 60.9\% mIoU \textit{without using any real data during training}. 
\subsection{Influence of Prompting Strategy}
Table \ref{tab:prompt_abl} shows the results of the resulting domain generalization when varying the prompts w.r.t.\ location, environment condition, and class-uniform sampling. 
Overall, the most important part of the text prompts seems to be the class-uniform sampling ($+\VECG{\varphi}^{\mathrm{CUS}}$) as it results in the highest domain generalization mIoU. 
But increasing the variation in each of the dimensions seems to have a positive effect, although the increase in performance is rather small for $\VECG{\varphi}^{\mathrm{loc}}\to\VECG{\varphi}^{\mathrm{loc+}}$ and $+\VECG{\varphi}^{\mathrm{con}}$. 
Increasing the variety of locations has the biggest effect on the performance on the Mapillary Vistas dataset \cite{Neuhold2017}, which consists of images from various locations all over the world. \par
However, increasing the number of conditions does not increase performance on MV or ACDC which are very diverse w.r.t.\ the conditions. This might be related to insufficiently realistic generation of such conditions by the diffusion model, or related to the fact that the HRDA \cite{hoyer2022hrda} method has difficulties to adapt to these conditions. 
Finally, we observe that using all text prompts does not improve over only class-uniform sampling. 
\subsection{Influence of UDA Approaches}
\begin{table}[t]
  \centering
  \renewcommand{\arraystretch}{.85}
  \setlength{\tabcolsep}{.1em}
  \caption{\textbf{Influence of DIDEX combined different UDA methods} on the generalization performance (mIoU(\%)).\textbf{Training} was performed on the synthetic \textbf{GTA5} ($\src\!=\!\gtavtrain$, upper part) or \textbf{SYNTHIA} ($\src\!=\!\synthiatrain$, lower part)  dataset. \textbf{Evaluation} is performed on various real-world \textbf{validation sets} ($\tgt\!=\!\datateststar$). Text prompts comprised the base prompt  $\VECG{\Phi}^{\mathrm{base}}$ and the CUS $\VECG{\varphi}^{\mathrm{CUS}}$. * indicates additional usage of the masked image consistency loss~\cite{hoyer2023mic}.}
  
 \extrarowheight=\aboverulesep
    \addtolength{\extrarowheight}{\belowrulesep}
    \aboverulesep=0pt
    \belowrulesep=0pt
    \begin{tabular}{cclhhhcc}
        \toprule
        
        &\multirow{1}{*}{\rot{\makecell[c]{\textbf{Enc.\hphantom{0}}}}}  & \multirow{2}{*}{\makecell[l] {\textbf{DIDEX + ...}}} &\multicolumn{5}{c}{\textbf{mIoU (\%) on}} \\
        \cline{4-8}
       &&&   $\csteststar$ & $\bddteststar$ & $\mvteststar$  &$\acdcteststar$&\raisebox{-4.0pt}{\shortstack{\textbf{DG} \\\textbf{mean}}}  \rule[-1.2ex]{0mm}{3.65ex} \\

        \midrule
        \parbox[t]{4mm}{\multirow{8}{*}{\rotatebox[origin=c]{90}{\small{{$\src$: \textbf{GTA5}}}}}} &\parbox[t]{2mm}{\multirow{4}{*}{\rotatebox[origin=c]{90}{\small{\network{\footnotesize ResNet}}}}}&DACS*~\cite{tranheden2021dacs}  & {$46.9 $} & {$40.0 $} & {${45.2} $} & {$31.9 $}&44.0\\
        &&SePiCo~\cite{xie2023sepico}   &44.9 & 36.4 &38.8 &30.4 &40.0 \\  
        &&DAFormer*~\cite{Hoyer2022daformer}  & {$50.4 $} & {\textbf{41.8}} & {$47.1$} & {$33.9 $}&46.4\\
        &&MIC~\cite{hoyer2023mic} & {$\textbf{52.4}$} & {$40.9$} & {$\textbf{49.2}$} &{$\textbf{36.1}$}&\textbf{47.5}\\
        \cmidrule{2-8}
        &\parbox[t]{2mm}{\multirow{4}{*}{\rotatebox[origin=c]{90}{\small{\network{\footnotesize Transformer}}}}} &DACS*~\cite{tranheden2021dacs}  & 52.0 & 49.0 & 53.7 &  41.9 & 51.8 \\
        &&SePiCo~\cite{xie2023sepico}   & 57.4 & 49.7 & 56.4 &  44.5 & 54.5 \\  
        &&DAFormer*~\cite{Hoyer2022daformer}   & 57.7 & \textbf{56.6} & 60.7 & 46.4 & 58.3 \\
        &&MIC~\cite{hoyer2023mic} & \textbf{62.0} & 54.3 & \textbf{63.0} &\textbf{50.1}  & \textbf{59.7}   \\
              
        \midrule[1.1pt]
        \parbox[t]{4mm}{\multirow{8}{*}{\rotatebox[origin=c]{90}{\small{{$\src$: \textbf{SYNTHIA}}}}}} &\parbox[t]{2mm}{\multirow{4}{*}{\rotatebox[origin=c]{90}{\small{\network{\footnotesize ResNet}}}}}&DACS*~\cite{tranheden2021dacs}  & {$47.5$} & {$38.6 $} & {$41.5 $} &{$30.1 $}&42.5\\
        &&SePiCo~\cite{xie2023sepico}&   43.3 & 32.6 & 40.6 &  27.7 & 38.8 \\
        &&DAFormer*~\cite{Hoyer2022daformer}  & {$49.8 $} & {$40.0 $} & {$45.5 $} &\textbf{33.7}&45.1\\
        &&MIC~\cite{hoyer2023mic} & \textbf{53.1 }& \textbf{41.8} &\textbf{50.3}  & 33.3&\textbf{48.4}\\
        \cmidrule{2-8}
         &\parbox[t]{2mm}{\multirow{4}{*}{\rotatebox[origin=c]{90}{\small{\network{\footnotesize Transformer}}}}}&DACS*~\cite{tranheden2021dacs}&   52.1 & 38.4 & 48.0 &  36.4 & 46.2 \\
        &&SePiCo~\cite{xie2023sepico}&   54.4 & 45.5 & 52.3 &  37.7 & 50.7 \\
        &&DAFormer*~\cite{Hoyer2022daformer}&   53.3 & 44.5 & 52.3 & 38.6 & 50.0 \\
        &&MIC~\cite{hoyer2023mic} &   \textbf{59.8} & \textbf{47.4} & \textbf{59.5} &\textbf{43.5}  & \textbf{55.6} \\
        \bottomrule

    \end{tabular}

  \label{tab:ablation_uda_methods} 
  \vspace{-4.5mm}
\end{table}
Table \ref{tab:ablation_uda_methods} shows the influence of the UDA approach that is used for the generalization by adaptation step to the pseudo-target domain. Recent UDA methods, such as DAFormer~\cite{Hoyer2022daformer} and MIC~\cite{hoyer2023mic} adapt better to the pseudo-target domain and thus generalize better across domains. It should be noted that even comparably simple UDA methods such as DACS \cite{tranheden2021dacs} obtain a high domain generalization and outperform previous SOTA methods for SYNTHIA as the source dataset.
However, SePiCo \cite{xie2023sepico} with a ResNet-101 backbone performs worse than the other methods which might be caused by the lack of hyperparameter optimization. 
\subsection{Influence of Image Quantity \& Consistency}
\begin{figure}[t]
    \centering
    \includegraphics[width=1\linewidth]{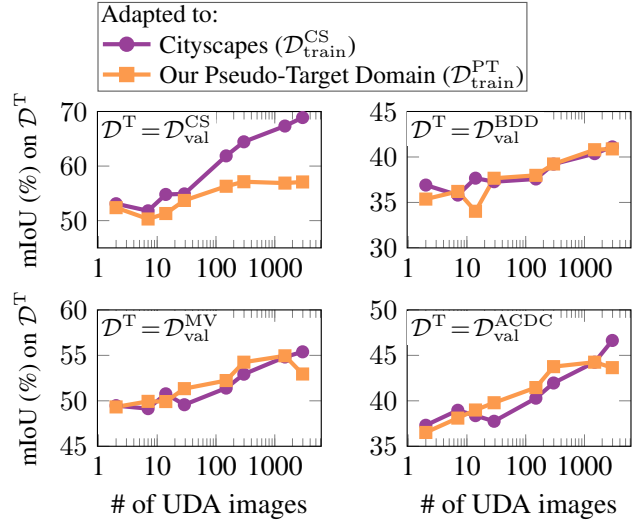}
    \caption{\textbf{Influence of the \# of UDA images} (target domain) on the generalization performance (mIoU (\%)). \textbf{Training} was performed on the synthetic \textbf{GTA5} ($\src\!=\!\gtavtrain$) dataset. \textbf{Evaluation} is performed on various real-world \textbf{validation sets} ($\tgt\!=\!\datateststar$). DAFormer~\cite{Hoyer2022daformer} was used for UDA and text prompts comprised only the base prompt  $\VECG{\Phi}^{\mathrm{base}}$ and the CUS block $\VECG{\varphi}^{\mathrm{CUS}}$.}
    \label{fig:miou_graph}
\end{figure}

In Figure \ref{fig:miou_graph}, we show results of randomly sampled subsets of either Citscapes (purple) or our pseudo-target domain (orange) used in the UDA approach. For each step, we applied DAFormer \cite{Hoyer2022daformer} for the adaptation from GTA5 to the subset.
We can observe that the model adapted to Cityscapes is on par with the pseudo-target domain model on Cityscapes up until a subset size of 29 UDA images, but gets increasingly better with more sampled images. 
This divergence of performance does, however, not occur for the other domains (BDD, MV, ACDC), where the models are mostly on par with each other. 
This indicates that the increase in performance on Cityscapes is caused by a specialization of the model to the target domain, but does not increase its generalization capabilities relative to the model adapted to the pseudo-target domain.
It further indicates that the images in the pseudo-target domain are of a similar "functional quality" for generalizing to other real-world domains as the real-world Cityscapes data.\par
In Table \ref{tab:ablation_semantic_consistency}, the influence of different ways to constrain the image generation process is shown. We observe that \network{ControlNet} \cite{zhang2023adding} constrained with semantics provides a small improvement over \network{SD2.0}~\cite{Rombach2022Stable} and can be a promising future research direction.   
\begin{table}[t]
  \centering
  \renewcommand{\arraystretch}{.8}
  \setlength{\tabcolsep}{.05em}
  \caption{\textbf{Influence of the diffusion model $\VECG{\Upsilon}$ architecture} and \textbf{contraints $\VEC{C}(\mathbf{x}_n)$} on the generalization performance (mIoU (\%)). \textbf{Training} was performed on the synthetic \textbf{GTA5} ($\src\!=\!\gtavtrain$) dataset. \textbf{Evaluation} is performed on various real-world \textbf{validation sets} ($\tgt\!=\!\datateststar$). UDA was performed with HRDA \cite{hoyer2022hrda} for \network{SD2.0}, MIC~\cite{hoyer2023mic} for the others, and a full prompt $\VECG{\Omega}_n$.}
  
 \extrarowheight=\aboverulesep
    \addtolength{\extrarowheight}{\belowrulesep}
    \aboverulesep=0pt
    \belowrulesep=0pt
    \begin{tabular}{llhhhcc}
        \toprule
        
        \multirow{2}{*}{\makecell[l] {\textbf{Diffusion}\\ \textbf{Model} $\VECG{\Upsilon}$}}&$\VEC{C}(\mathbf{x}_n)$&\multicolumn{5}{c}{\textbf{mIoU (\%) on}} \\
        \cmidrule{3-7}
       & &  $\csteststar$ & $\bddteststar$ & $\mvteststar$  &$\acdcteststar$&\raisebox{-4.0pt}{\shortstack{\textbf{DG} \\\textbf{mean}}}  \rule[-1.2ex]{0mm}{3.65ex} \\

        \midrule
       {\texttt{SD2.0}~\cite{Rombach2022Stable}}&Depth  & 58.8  & 52.7 & 63.2 & 47.4&58.3\\
        \texttt{ControlNet}~\cite{zhang2023adding} &Depth& 58.1 & \textbf{54.0} & 63.7 &46.7&58.6\\
        \texttt{ControlNet}~\cite{zhang2023adding}& Seg.& \textbf{58.9}& 53.9 & \textbf{64.1} & \textbf{49.0}&\textbf{59.0}\\
              
        \bottomrule

    \end{tabular}

  \label{tab:ablation_semantic_consistency} 
  \vspace{-4.5mm}
\end{table}

\section{Conclusions}
\label{sec:conclusions}
We introduced a novel diffusion-based domain extension (DIDEX) method for domain generalization which utilizes the generative capabilities of diffusion models. We projected the problem of domain generalization to the problem of domain adaptation which enabled us to utilize powerful adaptation methods for domain generalization. Our diffusion-based domain extension outperforms previous state-of-the-art methods by a large margin across datasets and architectures; with GTA5 as the source dataset by 3.8\% abs.\ mIoU and with SYNTHIA even by 11.8\% abs.\ mIoU on average.
Our text prompt ablation study has shown that information about present classes is beneficial for the pseudo-target domain. A remarkable result is that the functional quality of the diffusion generated data for the purpose of domain generalization is comparable to real Cityscapes data, highlighting the potential of using diffusion models for domain generalization.

\section*{Acknowledgment}
This work was supported by the German Federal Ministry for Economic Affairs and Climate Action within the project  “KI Wissen – Entwicklung von Methoden für die Einbindung von Wissen in maschinelles Lernen” and the SynthBAD project.

{\small
\bibliographystyle{ieee_fullname}
\bibliography{main}
}
\clearpage
\newpage
\clearpage
\setcounter{section}{0}
\section*{Supplementary Material}
 In this supplementary material we discuss limitations and ethical implications of our method. Additionally, we provide more samples including their segmentation masks for our pseudo-target domain. 
\renewcommand{\thesubsection}{\Alph{subsection}}
\subsection{Discussion of Limitations}
If data of a real domain is available, then the performance of current unsupervised domain adaptation (UDA) methods is still better than that of domain generalization (DG) method. 
However, given that UDA methods employ real-world data, and we only employ images that are generated by a diffusion model without accessing any real data, a comparison of these methods would not be adequate.
As shown in Figure \ref{fig:data-vis-main} and Figures~\ref{fig:data-vis-gta-controlnet}, ~\ref{fig:data-vis-synthia-controlnet}, ~\ref{fig:data-vis-gta-sd2}, and ~\ref{fig:data-vis-synthia-sd2} the text prompt has a crucial impact on the content of the generated image. We utilize a novel, systematic and modular text prompt strategy to obtain a diverse pseudo-target domain and validate it in Table \ref{tab:prompt_abl}. However, we believe that there is a strong potential to improve the prompt design and thereby the quality and diversity of the generated content. One idea might be automated prompt generation. Further details are left for future work. \par
Although our method does not require any data from any target domain, it should be noted that the employed diffusion models were naturally trained with real data. 
\subsection{Discussion of Ethical Implications}
The application of DIDEX includes the generation of data with diffusion models. Biases in this generated data have to be carefully considered since they are dependent on how the employed diffusion models were trained. Additional biases can be caused by the prompt generation, particularly, by the location string. When different locations shall be represented this may lead to ethnic biases, which was not within the scope of this paper. Biases in the employed real datasets also have to be considered when assessing the generalization performance. Of our four real datasets two (Cityscapes \cite{Cordts2016} and ACDC\cite{Sakaridis2021acdc}) were collected in Central Europe. BDD100k \cite{Yu2019} contains images from the USA, and only Mapillary Vistas \cite{Neuhold2017} represents multiple regions from all over the world. For real-world applications, especially when safety-critical, these biases have to be considered.\\ 
Although we have developed our method for applications such as automated driving or robotics, there is of course also the possibility of unintended use, e.g., surveillance or military applications. However, this is a general problem of methods that aim to make computer vision more robust.
\subsection{Further Example Images}
In the following we will provide more example images from both employed diffusion models with their respective prompt $\VECG{\Phi}_n$ in Figures~\ref{fig:data-vis-gta-controlnet}, ~\ref{fig:data-vis-synthia-controlnet}, ~\ref{fig:data-vis-gta-sd2}, and ~\ref{fig:data-vis-synthia-sd2}.
We display the text prompt at the top of each respective row and the input image $\mathbf{x}_n^{\mathrm{S}}$ at the left of the row. 
The image $\mathbf{x}_n^{\mathrm{PT}}$ is generated based on the prompt displayed in the middle. Additionally it is constrained by a depth estimation which is not shown.  
The segmentation result of the generated image is displayed at the right. 
Our best domain-generalizing segmentation model $\VEC{M}^{\mathrm{DG}}$ was used to create the prediction $\VEC{M}^\mathrm{DG}(\mathbf{x}_n^{\mathrm{PT}})$. 
Similar to Figure \ref{fig:data-vis-main} we can observe that generated images show semantic and structural inconsistencies. As expected, \network{ControlNet} \cite{zhang2023adding} shows a higher degree of consistency than \network{SD2.0} \cite{Rombach2022Stable}. However, even with strong inconsistencies the model adapts well to our pseudo-target domain as the segmentation predictions reveal. Please note that since we do not have labels for the pseudo-target domain we cannot calculate the corresponding mIoU. 
\newpage
\begin{figure*}
    \centering
    \includegraphics[width=\linewidth]{fig/supplement_samples_controlnetv2.tex}
    \caption{\textbf{Samples from the pseudo-target domain $\mathcal{D}^{\mathrm{PT}}$} generated with \network{ControlNet} \cite{zhang2023adding} with depth constraints and based on $\mathcal{D}^{\mathrm{S}}=\gtavtrain$ images. The predictions were obtained with our best domain-generalizing model $\VEC{M}^{\mathrm{DG}}$.}
    \label{fig:data-vis-gta-controlnet}
\end{figure*}

\begin{figure*}
    \centering
    \includegraphics[width=\linewidth]{fig/supplement_synthia_visv2.tex}
    \caption{\textbf{Samples from the pseudo-target domain $\mathcal{D}^{\mathrm{PT}}$} generated with \network{ControlNet} \cite{zhang2023adding} with depth constraints and based on $\mathcal{D}^{\mathrm{S}}=\synthiatrain$ images. The predictions were obtained with our best domain-generalizing model $\VEC{M}^{\mathrm{DG}}$.}
    \label{fig:data-vis-synthia-controlnet}
\end{figure*}

\begin{figure*}
    \centering
    \includegraphics[width=\linewidth]{fig/supplement_samples_stable_diff2_gta.tex}
    \caption{\textbf{Samples from the pseudo-target domain $\mathcal{D}^{\mathrm{PT}}$} generated with \network{SD2.0} \cite{Rombach2022Stable} with depth constraints and based on $\mathcal{D}^{\mathrm{S}}=\gtavtrain$ images. The predictions were obtained with our best domain-generalizing model $\VEC{M}^{\mathrm{DG}}$.}
    \label{fig:data-vis-gta-sd2}
\end{figure*}

\begin{figure*}
    \centering
    \includegraphics[width=\linewidth]{fig/supplement_samples_stable_diff2_synthia.tex}
    \caption{\textbf{Samples from the pseudo-target domain $\mathcal{D}^{\mathrm{PT}}$} generated with \network{SD2.0} \cite{Rombach2022Stable} with depth constraints and based on $\mathcal{D}^{\mathrm{S}}=\synthiatrain$ images. The predictions were obtained with our best domain-generalizing model $\VEC{M}^{\mathrm{DG}}$.}
    \label{fig:data-vis-synthia-sd2}
\end{figure*}

\end{document}


\graphicspath{{./fig}{../fig}}
\makeatletter
  \newcommand\tinyv{\@setfontsize\tinyv{2pt}{6}}
\makeatother
\def\x{5.1cm}  
\def\y{3.6cm} 
\def\shift{4.5cm}  

\begin{tikzpicture}[
 ] 
 \setlength{\baselineskip}{4pt}
\node at (0,0) (frame1)
    {\includegraphics[width=5cm]{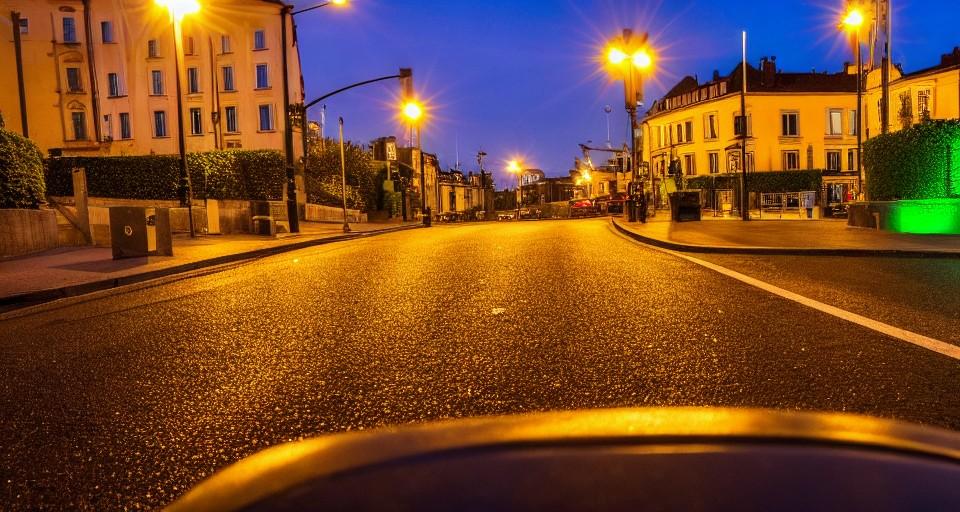}};
\node[left of= frame1, node distance=5.0cm] (frame4)
    {\includegraphics[width=4.8cm]{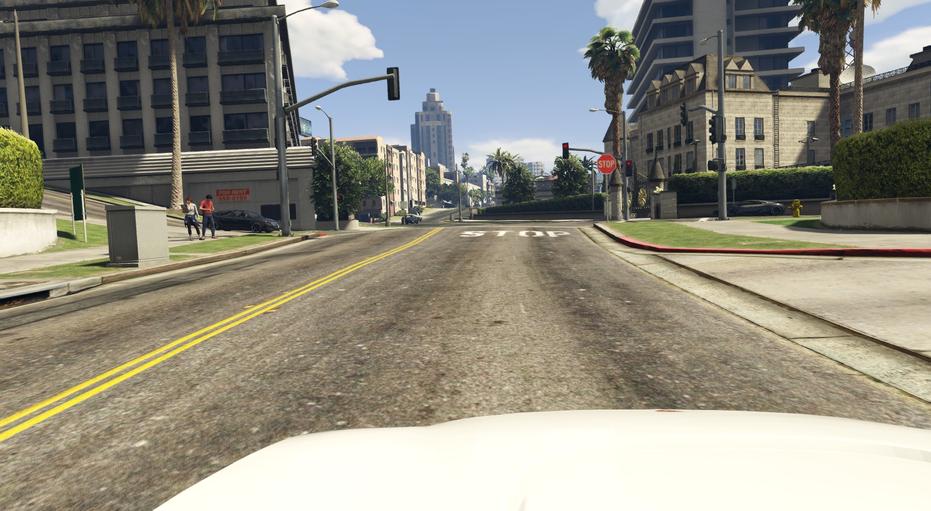}};
\node[right of= frame1, node distance=\x] (frame7)
    {\includegraphics[width=5cm]{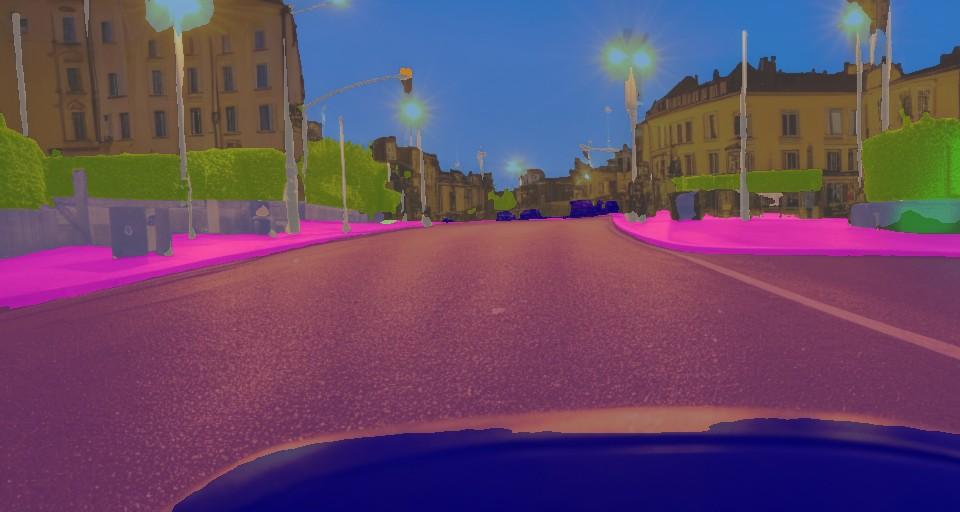}};
\node [above of=frame1, node distance=1.8cm, text width=14cm, execute at begin node=\setlength{\baselineskip}{4pt}, draw, color=black](tex1){\footnotesize $\VECG{\Phi}_n\!=$"\textit{A high quality photo; Europe; rider, road, sidewalk, building, wall, fence, pole, traffic light, traffic sign, vegetation, terrain, sky, person, car, evening}"} ;

\node[below of=frame1, node distance=\y] (frame8)
    {\includegraphics[width=5cm]{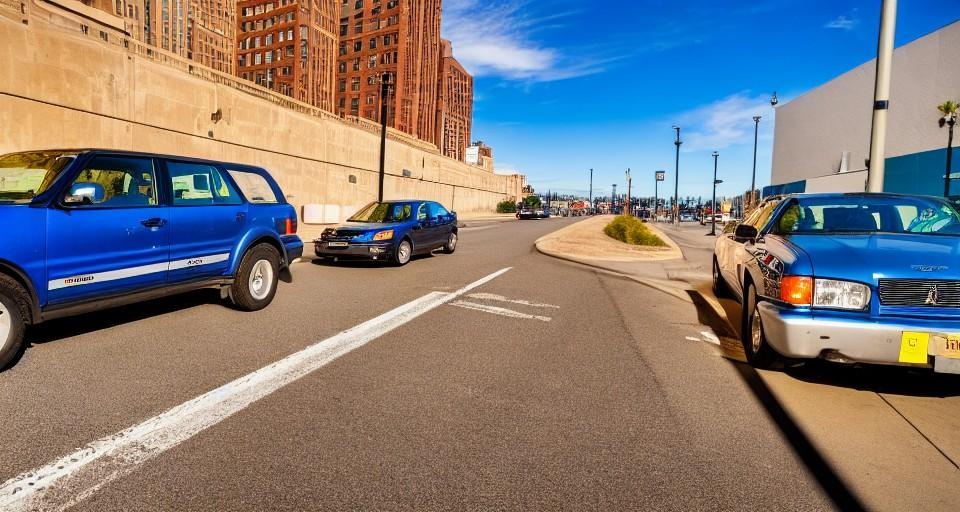}};
\node[below of= frame4, node distance=\y] (frame9)
    {\includegraphics[width=4.8cm]{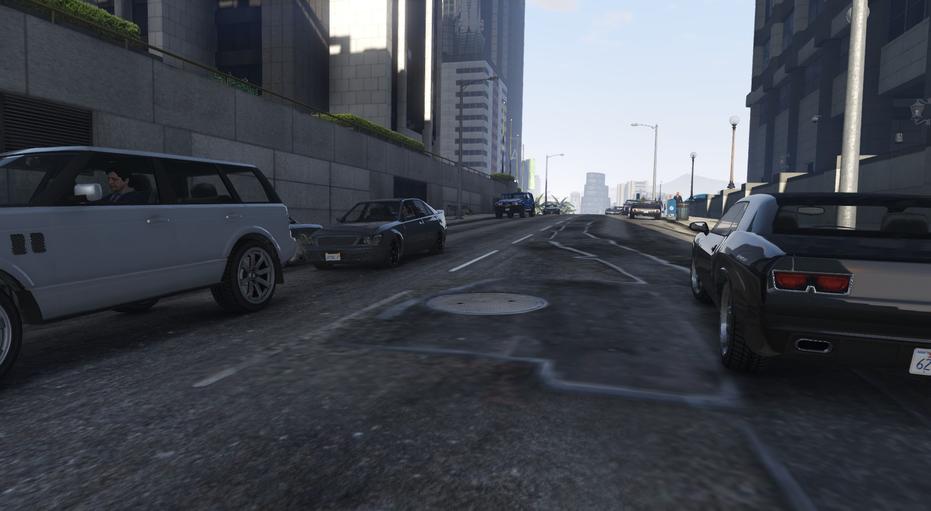}};
\node[below of= frame7, node distance=\y] (frame10)
    {\includegraphics[width=5cm]{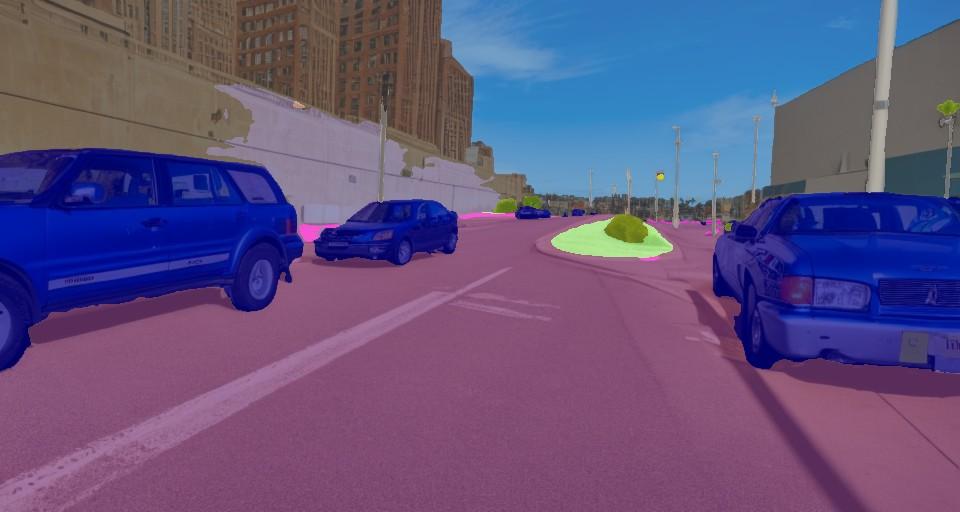}};
\node [above of=frame8, node distance=1.8cm, text width=14cm, execute at begin node=\setlength{\baselineskip}{4pt}, draw, color=black](tex1){\footnotesize $\VECG{\Phi}_n\!=$"\textit{A high quality photo; USA; city, bicycle, road, sidewalk, building, wall, fence, pole, traffic sign, vegetation, terrain, sky, person, car, sunny}"} ;

\node[below of=frame8, node distance=\y] (frame11)
    {\includegraphics[width=5cm]{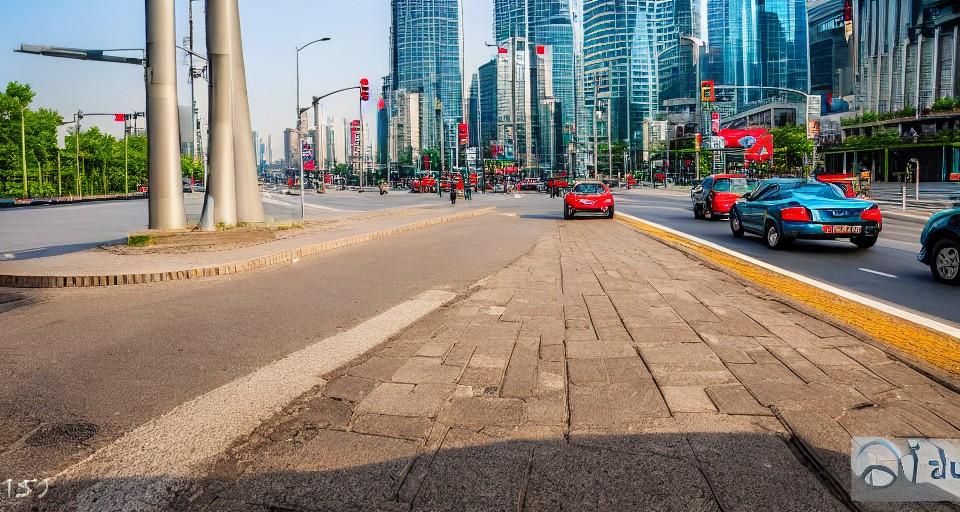}};
\node[below of= frame9, node distance=\y] (frame12)
    {\includegraphics[width=4.8cm]{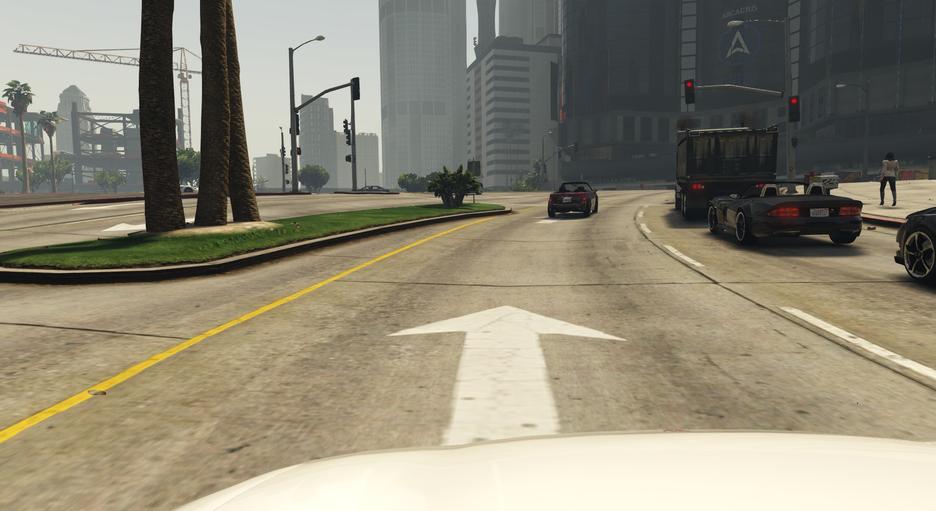}};
\node[below of= frame10, node distance=\y] (frame13)
    {\includegraphics[width=5cm]{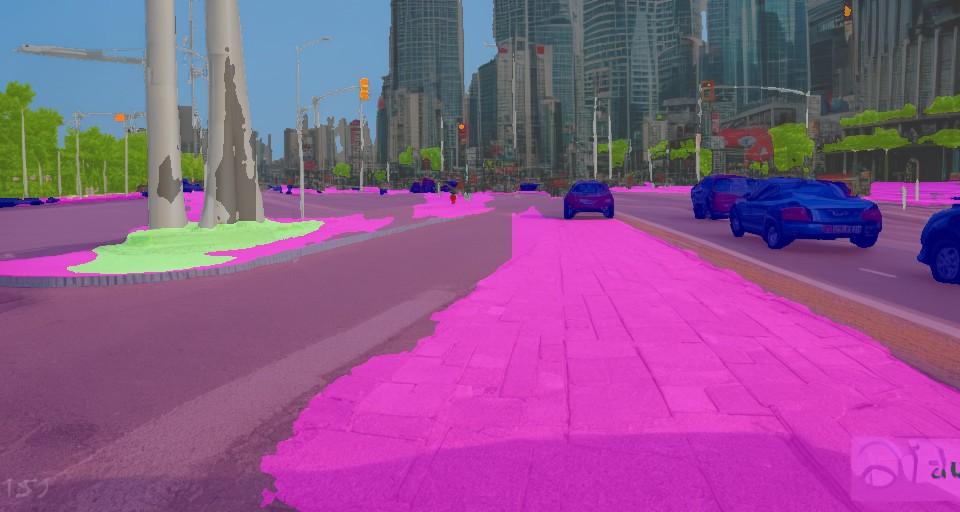}};
\node [above of=frame11, node distance=1.8cm, text width=14cm, execute at begin node=\setlength{\baselineskip}{4pt}, draw, color=black](tex1){\footnotesize $\VECG{\Phi}_n\!=$"\textit{A high quality photo; China; city, bicycle, road, sidewalk, building, wall, fence, pole, traffic light, traffic sign, vegetation, terrain, sky, person, car, truck, summer}"} ;

\node[below of=frame11, node distance=\y] (frame14)
    {\includegraphics[width=5cm]{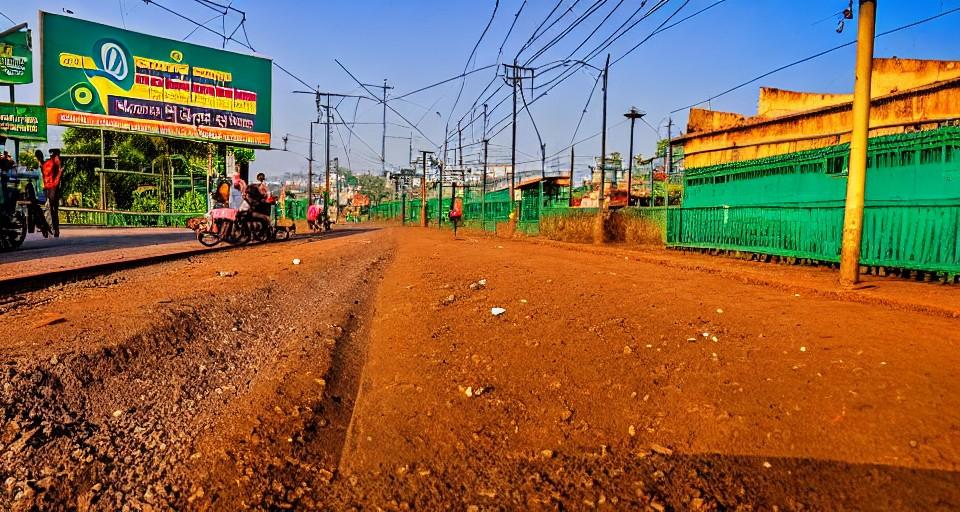}};
\node[below of= frame12, node distance=\y] (frame15)
    {\includegraphics[width=4.8cm]{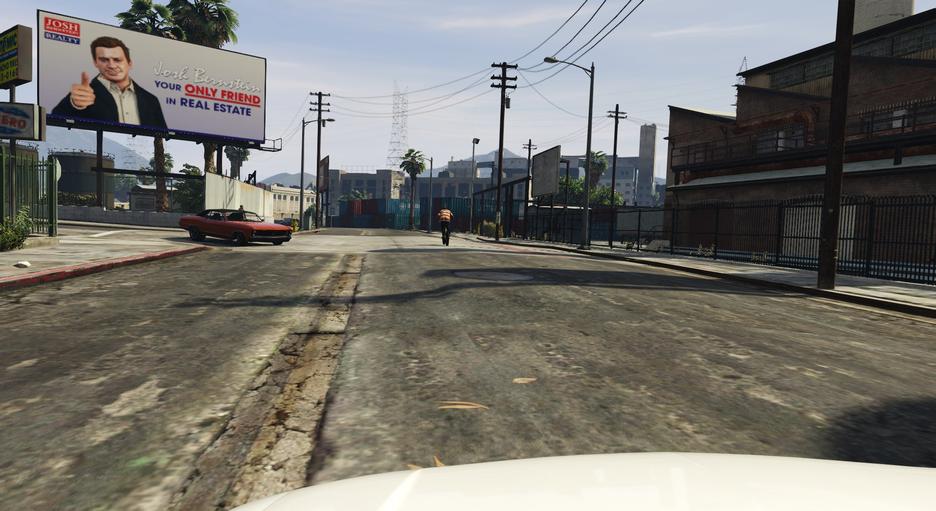}};
\node[below of= frame13, node distance=\y] (frame16)
    {\includegraphics[width=5cm]{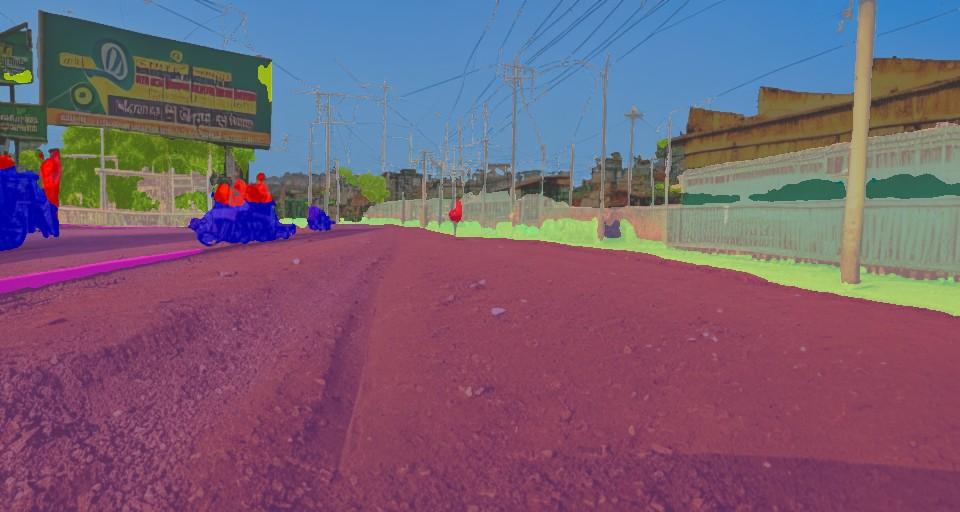}};
\node [above of=frame14, node distance=1.8cm, text width=14cm, execute at begin node=\setlength{\baselineskip}{4pt}, draw, color=black](tex1){\footnotesize $\VECG{\Phi}_n\!=$"\textit{A high quality photo; India; city, train, road, sidewalk, building, wall, fence, pole, vegetation, terrain, sky, person, rider, car, bicycle, harsh light}"} ;

\node[below of=frame14, node distance=\y] (frame17)
    {\includegraphics[width=5cm]{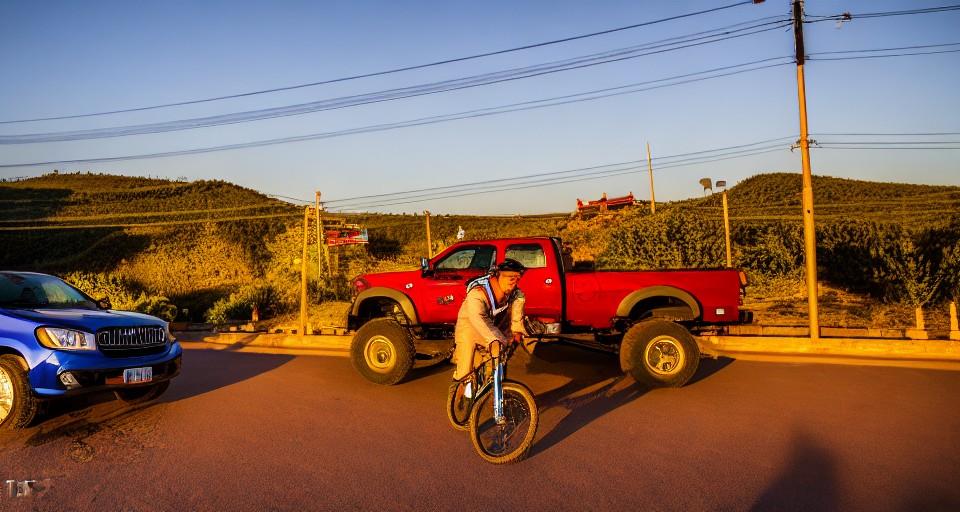}};
\node[below of= frame15, node distance=\y] (frame18)
    {\includegraphics[width=4.8cm]{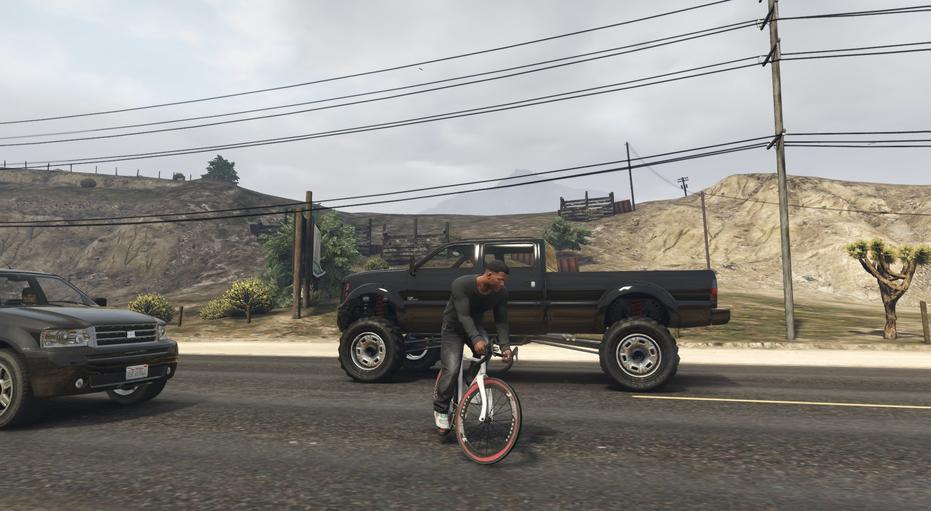}};
\node[below of= frame16, node distance=\y] (frame19)
    {\includegraphics[width=5cm]{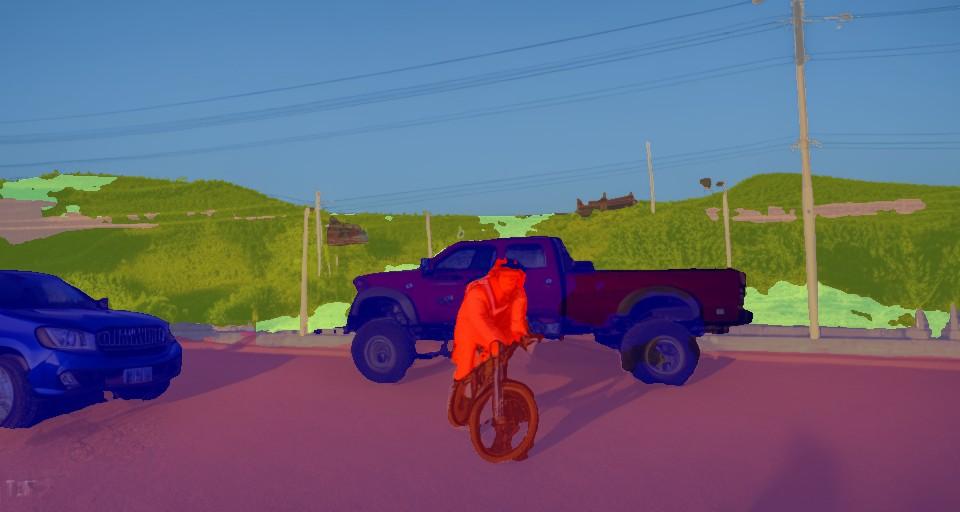}};
\node [above of=frame17, node distance=1.8cm, text width=14cm, execute at begin node=\setlength{\baselineskip}{4pt}, draw, color=black](tex1){\footnotesize $\VECG{\Phi}_n\!=$"\textit{A high quality photo; China; rider, road, building, fence, pole, vegetation, terrain, sky, person, car, truck, evening}"} ;

 \node [above of=frame4, node distance=2.8cm](tex1){ \small\textsf{\textbf{GTA5 Input $\mathbf{x}_n^{\mathrm{S}}$}}} ;
 \node [above of=frame1, node distance=2.8cm](tex1){\small\textsf{\textbf{Diffusion Sample $\mathbf{x}_n^{\mathrm{PT}}$}}} ;
 \node [above of=frame7, node distance=2.8cm](tex1){ \small\textsf{\textbf{Prediction $\VEC{M}^\mathrm{DG}(\mathbf{x}_n^{\mathrm{PT}})$}}} ;
\end{tikzpicture}